\documentclass[10pt,twocolumn,letterpaper]{article}

\usepackage{iccv}
\usepackage{times}
\usepackage{epsfig}
\usepackage{graphicx}
\usepackage{amsmath}
\usepackage{amssymb}

\usepackage{amsthm}
\usepackage{appendix}
\usepackage{xcolor, colortbl}
\definecolor{Gray}{gray}{0.7}
\definecolor{Gray1}{gray}{0.9}
\usepackage[skip=3pt,font=small]{caption}

\usepackage{pifont}
\newcommand{\cmark}{\ding{51}}%
\usepackage{subfig}
\usepackage{multirow}
\usepackage[ruled,vlined,linesnumbered]{algorithm2e}

\usepackage{subfig}

\usepackage[pagebackref=true,breaklinks=true,letterpaper=true,colorlinks,bookmarks=false]{hyperref}

\iccvfinalcopy 

\def\iccvPaperID{2215} 

\ificcvfinal\fi

\begin{document}

\title{Learning Compatible Embeddings}

\author{Qiang Meng \quad Chixiang Zhang \quad Xiaoqiang Xu \quad Feng Zhou\\
  Algorithm Research, AiBee Inc.\\
  {\tt\small \{qmeng, cxzhang, xiaoqiangxu, fzhou\}@aibee.com}
}

\maketitle
\ificcvfinal\thispagestyle{empty}\fi

\begin{abstract}
  Achieving backward compatibility when rolling out new models can highly reduce costs or even bypass feature re-encoding of existing gallery images for in-production visual retrieval systems.
  Previous related works usually leverage losses used in knowledge distillation which can cause performance degradations or not guarantee compatibility.
  To address these issues, we propose a general framework called Learning Compatible Embeddings (LCE) which is applicable for both cross model compatibility and compatible training in direct/forward/backward manners.
  Our compatibility is achieved by aligning class centers between models directly or via a transformation, and restricting more compact intra-class distributions for the new model.
  Experiments are conducted in extensive scenarios such as changes of training dataset, loss functions, network architectures as well as feature dimensions, and demonstrate that LCE efficiently enables model compatibility with marginal sacrifices of accuracies.
  The code will be available at \url{https://github.com/IrvingMeng/LCE}.
\end{abstract}

\section{Introduction}

Visual search or retrieval systems~\cite{scheirer2012toward, scheirer2014probability} are widely used in many real-world applications such as face recognition~\cite{taigman2014deepface, sun2015deepid3, Ranjan2017, meng2021poseface, meng2021magface}, person re-identification~\cite{Sun2018Beyond, Hermans2017In, ristani2018features}, car re-identification~\cite{khan2019survey} and image retrieval~\cite{bendale2016towards, gordo2016deep}.
To obtain steady improvement, models would be occasionally upgraded by training on larger or cleaner datasets, adopting more powerful network structures and training losses, or applying techniques like network architecture search~\cite{zoph2016neural,xu2020searching}, knowledge distillation~\cite{hinton2015distilling} and network pruning~\cite{liu2018rethinking, frankle2018lottery}.
However, to harvest the benefits of new models, a process known as ``backfilling'' or ``re-indexing''~\cite{shen2020towards} is indispensable to re-encode all images in the gallery set to recreate clusters.
This process could be impractical to execute when there are limited computational resources for re-encoding, or original images are legally forbidden to be preserved without user authorization. 
Model compatibility techniques, which can heavily reduce costs of or even bypass the process, therefore are of great practical values.

\begin{figure}[t]
  \centering
  \includegraphics[width=0.48\textwidth]{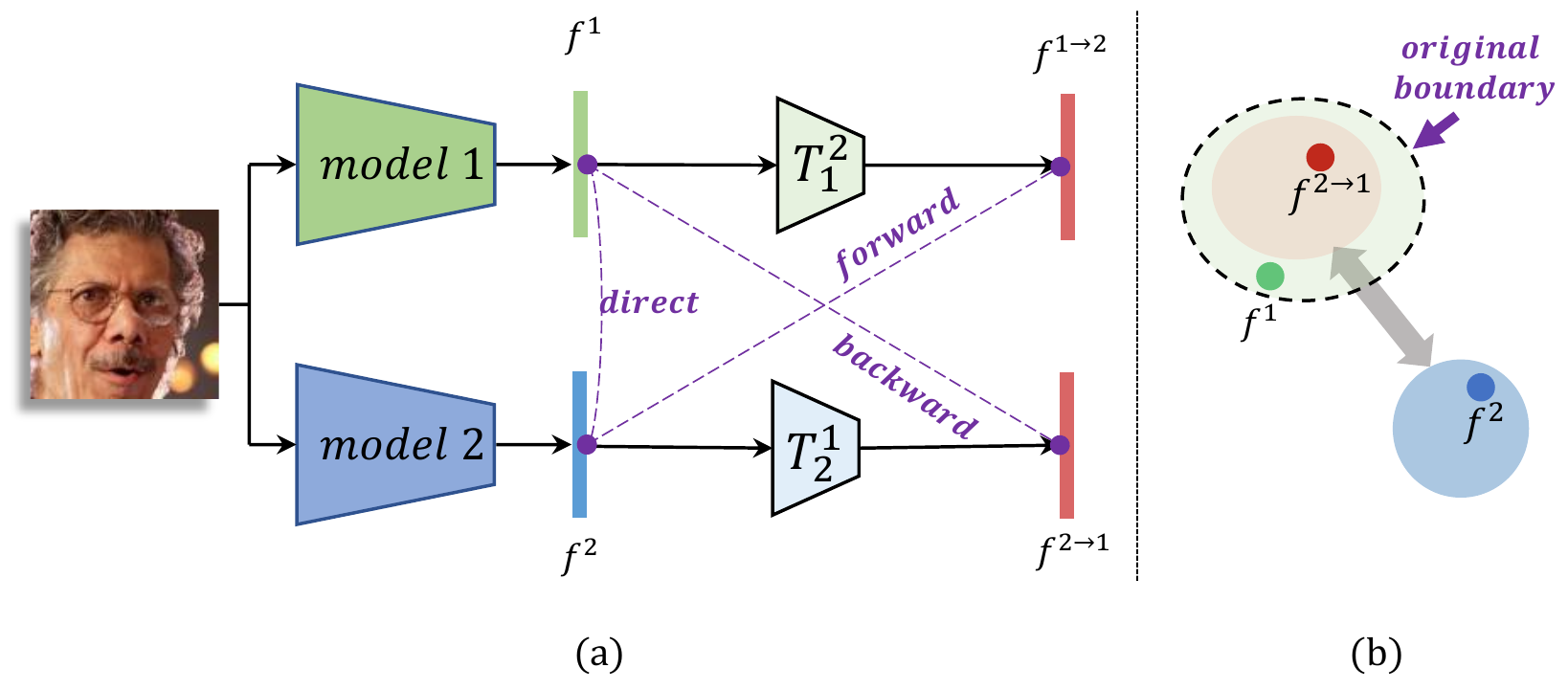}
  \caption{
    An overview of model compatibility problems.
    (a) Model compatibility matches model 1 and model 2 through three optional directions.
    Features $f^1$, $f^2$ from two models can make a direct comparison, noted as direct compatible method.
    With transformations $T_1^2$ and $T_2^1$ involved, we achieve either backward compatible by comparing $f^{2\rightarrow 1}$ with $f^1$, or forward compatible if compare $f^{1\rightarrow 2}$ and $f^2$.
    (b) Our compatibility is achieved by aligning the class centers across models through a direct comparison (by setting $T$s to identity mappings) or via transformations, and restricting more compact intra-class distributions for the new model.
    \textbf{Best viewed in color.}
  }\label{fig:intro}
\end{figure}

Because of the multiplicity of application scenarios in visual search/retrieval fields, as well as compatibility directions to implement, model compatibility has never been employed via a general, unified set of patterns.
Viewing in the perspective of application scenarios, there are two types of model compatibility methods.
The first type is called cross model compatibility (CMC) whose goal is to find compatible mappings between the previous and upgraded models, where the upgraded models already exist.
As model performance may degrade if compatibility is considered, CMC serves well for the scenarios that model performances are more valuable than compatibility.
The second type is compatible training (CT) which aims at upgrading models with compatibility constraints.
When bringing up against the scenarios where frequent iterations for online models are required, CT possesses an inherent feasibility to take charge.

The perspective of compatible directions categorizes model compatibility into three types: backward, forward and direct compatible methods (Fig.~\ref{fig:intro}a).
Direct compatible methods compare new features with old gallery sets directly, which is the most efficient as it completely prevents ``backfilling'' processes.
Backward compatible methods endow a backward transformation that maps new features into old feature spaces, and they are capable of waiving re-encoding large gallery sets.
Lightweight improvement for cumbersome model architectures is another potential scenario for backward compatible methods since old models spawn more separable feature spaces than new models in the circumstances.
Forward compatible methods utilize a forward transformation that maps old features into new feature spaces, which is aimed at upgrading small gallery sets, especially when new models have much better performances than old ones.

Despite its great practical values, model compatibility is a relatively unexplored research area.
Few efforts, R$^3$AN~\cite{chen2019r3}, RBT~\cite{wang2020unified} (for CMC) and BCT~\cite{shen2020towards} (for CT), are proposed and reach comparable results.
However, these three works are only appropriate for limited application scenarios or compatible directions.
Moreover, their model performances and compatibility are still arduous to be guaranteed, not only because of the further intensified nature of incompatibility for two different feature spaces when upgraded models dramatically changing training factors (\eg, backbones, losses, training datasets, or even settings of optimizers), but also as a result of the point-wise losses they utilize,
which align each feature pair and therefore prevent learning of more discriminative features.


To address the above issues, we propose a general framework called LCE for model compatibility as shown in Fig.~\ref{fig:intro}.
Model 1 represents the target compatible model.
Model 2 is pre-existed for CMC while trained for CT.
Transformations $T^1, T^2$ map features to another spaces and enable direct/backward/forward compatibility.
Our compatibility is achieved by aligning feature classes across models and restricting the mapped features to be distributed within the original boundaries (\ie, more compact intra-class distributions).
By this means, the mapped feature $f^{2\rightarrow 1}$ is distributed inside the correct class in the original feature space and that makes features comparable.

Besides the compatibility, our framework also benefits the learning of more discriminative features in the following aspects:
(a) Our method works in a point-to-set manner instead of employing point-wise constraints.
As shown in Fig.~\ref{fig:intro}b, features from same instance are not restricted to be distributed closely.
(b) By the restriction of the original boundary, intra-class distributions are more compact compared to those from the old model.
(c) The introduced transformation $T$ relaxes the requirement for consistent inter-class distributions, as inter-class distributions may vary a lot across models (\eg, ResNet100~\cite{he2016deep} can separate features more easily than MobileFace~\cite{chen2018mobilefacenets}).
Overall, we make the following contributions:
\begin{itemize}
  \itemsep0em
  \item Compared to previous methods which work in a point-wise manner,
    we reformulate the model compatibility from the aspect of classes and highly decouple models.
    Specifically, we align feature classes across models and restrict the new-to-old mapped features to be distributed within the original boundaries.
    With the proposed point-to-set constraints and transformations, we achieve model compatibility with marginal sacrifices of accuracies.
  \item We unify the two model compatibility problems (\eg, CMC and CT) into a unified training framework called LCE, which can work in direct/backward/forward manners.
    Extensive experiments and ablation study are conducted on both problems with various factors such as network structures and losses.
    The proposed method achieves noteworthy results compared to current state-of-the-arts.
  \end{itemize}

\section{Related Works}
Model compatibility is essentially transferring knowledge across models.
This section reviews related works in fields of transfer learning, knowledge distillation and model compatibility, from the perspective of feature embeddings.

\subsection{Feature Representation Transfer Learning}
Feature representation transfer learning aims at transforming each original feature into a new feature representation for knowledge transfer~\cite{zhuang2020comprehensive}.
Pan \etal~\cite{pan2008transfer} propose to learn a low-dimensional latent feature space where the distributions between the source and target domain data are the same or close to each other.
JDA~\cite{long2013transfer} jointly adapts both the marginal distribution and conditional distribution and constructs new feature representation for substantial distribution difference.
JDA is further extended by utilizing the label and structure information~\cite{hou2016unsupervised}, clustering information~\cite{tahmoresnezhad2017visual}, various statistical and geometrical information~\cite{zhang2017joint} and balanced distributions~\cite{wang2017balanced}, \etc.
Feature augmentation based methods~\cite{daume2009frustratingly, kumar2010co, duan2012learning, li2013learning} transfer the original features by feature replication to augment datasets for target domain.
Even though is related to model compatibility, feature representation transfer learning focuses on transferring the knowledge contained in source domains to target domains instead of achieving comparable features.

\subsection{Knowledge Distillation}
Knowledge distillation (KD), which aims at transferring knowledge acquired in a teacher model to a student model, was first proposed by Hinton \etal~\cite{hinton2015distilling}.
There exists a great amount of knowledge sources and knowledge types in this field (see~\cite{gou2020knowledge, wang2021knowledge} for recent reviews).
Various knowledge sources are adopted in literature such as classification logits~\cite{hinton2015distilling, ba2013deep, mirzadeh2020improved}, hint layers~\cite{romero2014fitnets, zhou2018rocket, wangexclusivity, chen2020learning} and multi-layer groups~\cite{zagoruyko2016paying, yue2020mgd, kim2018paraphrasing}.
Knowledge types can be categorized into feature embeddings~\cite{romero2014fitnets, wangexclusivity}, feature maps~\cite{chen2020learning} and instance relationships~\cite{park2019relational, peng2019correlation}, \etc.
Despite the diverse categories, most works adopt  KL-divergence or $l_2$ distance as the loss to transfer knowledge.
For example, both DarkRank~\cite{chen2018darkrank} and Mirzadeh \etal~\cite{mirzadeh2020improved} learn to match softened logits between teacher model and student model using KL-divergence.
PKT~\cite{passalis2020probabilistic} proposes to align probability distributions in the feature spaces for representation/metric learning tasks.
RKD~\cite{park2019relational} introduces distance-wise and angle-wise distillation losses that mimic structural differences in relations.
CCKD~\cite{peng2019correlation} combines the two losses to ensure congruence between teacher and student models, in both instance-level and group-level.
Their experiments are conducted on classification tasks and metric learning tasks and achieve promising results.


\subsection{Model Compatibility}\label{sec:method_revisit}
 Model compatibility is to make features comparable across models and is of great practical values for visual retrieval systems.
This technique can heavily reduce computational costs of re-encoding gallery sets which is large-volume in modern applications but less studied in literatures.
For CMC whose goal is to find mappings between models,
R$^3$AN~\cite{chen2019r3} learns feature transformation by a process of reconstruction, representation and regression for face recognition.
RBT~\cite{wang2020unified} designs a light-weight residual bottleneck transformation module following the split-transform-merge strategy and achieves remarkable improvements compared to R$^3$AN in their experiments.
Their module is trained by a classification loss, an $l_2$ similarity loss between feature embeddings and a KL-divergence loss between logits in classifiers.
For CT, BCT~\cite{shen2020towards} introduces an influence loss and trains new features on both new and old classifiers, which is essentially aligning the logits.
As sharing the same high-level idea of transferring knowledge in KD, RBT~\cite{wang2020unified} and BCT~\cite{shen2020towards} both utilize similar ideas and losses to achieve their compatibility.

\begin{figure}
  \centering
  \includegraphics[width=0.4\textwidth]{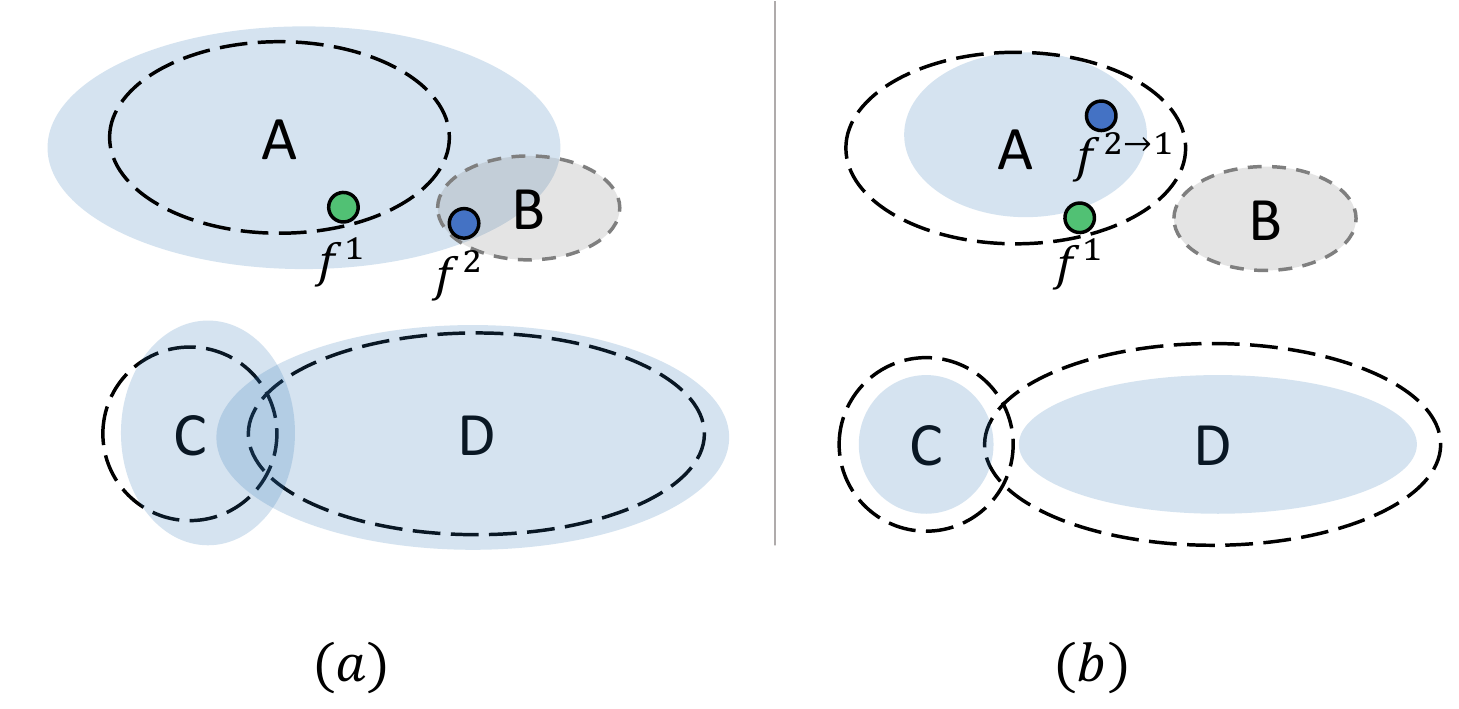}
  \caption{Model compatibility in an open-set system.
    $f^1, f^2$  are features from model 1, 2.
    Regions inside dash lines represent feature clusters from model 1 and the colored region are from  model 2.
    In this example, class A, C, D are from the training dataset.
    Class B is an absent class in training dataset but can appear during test.
    (a) Traditional methods can lead to incompatible feature spaces.
    (b) Feature distributions in our method.
    Here $f^{2\rightarrow 1}$ is the mapped feature, whose distribution is restricted inside the boundary of model 1.
  }\label{method:0}
\end{figure}

The main differences between KD and model compatibility mainly lie in two aspects.
The first difference is that the old model in KD is a teacher model and normally performs better than the student model.
For model compatibility, performances of upgraded models may decrease in some special cases such as light-weight structures or small feature dimensions introduced.
However, in most scenarios, the upgrade models have potentially better performances with more data involved and technological advances.
Directly distilling the exact logits or locations between each pair of old and new features, usually adopted by previous works, is too strict and could inherit bad inter-class distributions from old models to new ones.
For example, class C and D in the old feature space are overlapped in Fig.~\ref{method:0}a.
Directly aligning features prevents the new model from learning more discriminative features.

Another difference is that traditional losses used in KD cannot fully guarantee model compatibility, as intra-class boundaries are difficult to be constrained by traditional KD losses.
As shown in Fig.~\ref{method:0}a, $f^1, f^2$ have similar logits and a small $l_2$ distance during training because of the absence of class B.
Therefore, the conventional losses in KD perform weak penalties.
While during the testing phase, $f^2$ belongs to cluster B and thus is not compatible with the original feature $f^1$.

Our proposed LCE framework is capable of alleviating the vulnerabilities of traditional KD losses.
As Fig.~\ref{method:0}b depicts, our method not only forces more compact intra-class distributions (class A, C, D), but also renders more discriminative inter-class distributions (class C, D are separable with new features) for the new model.

\section{Methodology}

  \begin{figure*}
  \centering
  \includegraphics[width=0.85\textwidth]{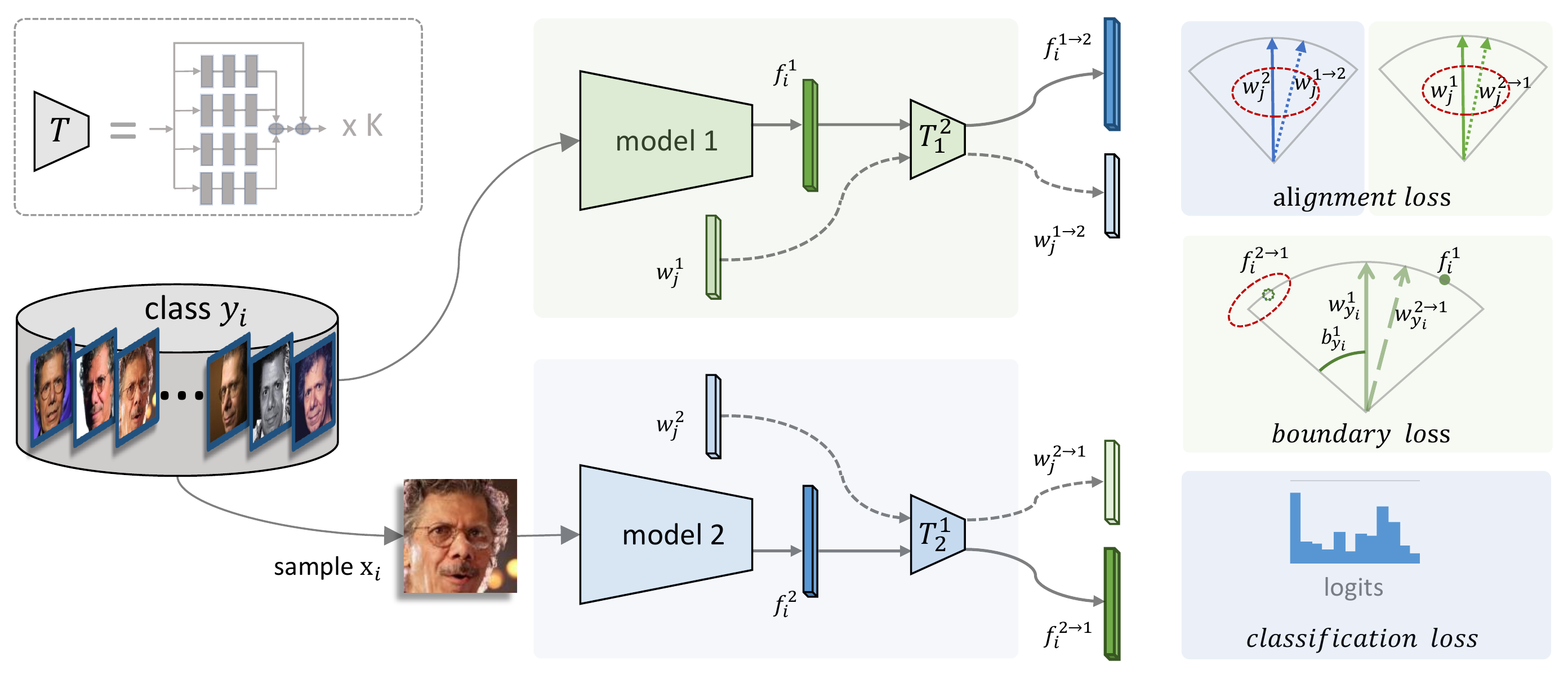}
  \caption{Overview of the proposed LCE method and the associated losses.
    We assume model 1 is the compatible target model and therefore its weights are fixed during the training.
    In the feature space of model 1, denote class $j$ has a maximum angle $b_j^1$   to its class center $w_j^1$.
    The model 2 can be pre-existed or to be trained based on the real situations, and transformations $T_1^2, T_2^1$ map embeddings $\{f_i\}_{i=1}^{N}$ and class centers $\{w_j\}_{j=1}^{n}$ from one feature space to another.
    Our compatibility is achieved by
    (1) an alignment loss which aligns class centers between two models,
    (2) a boundary loss which enforces the mapped $f_i^{2\rightarrow 1}$ to have more compact intra-class distributions
    (formulated as the angle between $f_i^{2\rightarrow 1}$ and $w_{y_i}^1$ is smaller than $b_{y_i}^1$ in our work)
    and (3) a classification loss. }\label{fig:method1}
\end{figure*}

Our LCE framework is developed following three rules:
(1) Compatibility should be ensured and the issues presented in Fig.~\ref{method:0}a are expected to be prevented whenever possible.
(2) The new model should be decoupled from the previous one as much as possible to reduce performance drops.
(3) The method should fit both CMC and CT problems.

\subsection{Criterions for Compatibility}
Suppose there are $N$ samples $\{x_i, y_i\}_{i=1}^N$ of $n$ classes and their embeddings from two models are $\{f_i^1, f_i^2\}_{i=1}^N$, respectively.
For $\forall i_1, \forall i_2 \in \{1, 2, \cdots N\}$ and $i_1 \neq i_2$, the point-wise compatibility criterion (\eg, BCT~\cite{shen2020towards}) is defined as 
\begin{equation}
  \label{eq:method_bctcompat}
  \small
  \text{(Point-wise) \quad  }
  \begin{aligned}
    d(f_{i_1}^2, f_{i_2}^1) \geq d(f_{i_1}^1, f_{i_2}^1),&\text{ if }   y_{i_1}\neq y_{i_2} ,\\
    d(f_{i_1}^2, f_{i_2}^1) \leq d(f_{i_1}^1, f_{i_2}^1),&\text{ if }   y_{i_1} = y_{i_2}.
  \end{aligned}
\end{equation}
Here $d$ measures the distances.
This criterion is too strict as it adds constraints to all pairs of samples (\ie, $N^2-N$ constraints) and leads to excluded solutions especially if models are trained with different networks/losses.

Our compatibility is defined differently.
To start with, we first map features to another feature space, which relaxes the requirement for direct comparisons.
We call the mapped features as $f_i^{1\rightarrow2}, f_i^{2\rightarrow 1}$.
Denote features of class $j$ from model 1  are distributed in a feature region $C_{j}^1$.
Our criterion for compatibility is simply defined as $f^{2\rightarrow 1}_i\in  C_{y_i}^1$ for all $i$,
which is from the perspective of point-to-set comparison.
Specifically, we restrict the $f_i^{2\rightarrow 1}$ to be distributed inside the corresponding original boundary as indicated in Fig.~\ref{method:0}b.
By this means, the new feature space regulates more constrictive intra-class boundaries and more diacritical inter-class distributions.

As the real feature region is unavailable (\ie, a feature region contains an infinity number of features while the estimated features are of a limited number), we estimate the set $C_{y_i}^1$ by the class center $w_{y_i}^1$ and its boundary $b_{y_i}^1$ (as shown in Fig.~\ref{fig:method1}, boundary loss).
Then the criterion is re-formulated by
\begin{equation}
  \label{eq:method_compat_new}
  \small
  \text{(Point-to-set)  \quad  }
  \begin{aligned}
    \theta(f^{2\rightarrow 1}_i, w_{y_i}^1) \leq b_{y_i}^1 \quad \forall i \in \{1,2, \cdots, N\}.
  \end{aligned}
\end{equation}
Here $\theta(w, f) = \arccos(\frac{w^Tf}{\|w\|\|f\|})$ which measures the angle between vectors $w$ and $f$.

\subsection{Transformation Module}
Besides the point-to-set compatibility, we further decouple models by transformations $T_1^2, T_2^1$ which relieve gaps between feature spaces preferred by various models.
The transformations $T_1^2, T_2^1$ map features/class centers from one feature space to another and mapped embeddings are denoted by superscripts $1\rightarrow 2, 2\rightarrow 1$.
Specifically, we have
\begin{equation}
  \label{eq:method_map}
  \small
  \begin{aligned}
    f^{2\rightarrow 1}_i = T_2^1(f_i^2), \   f^{1\rightarrow 2}_i = T_1^2(f_i^1), \quad  \forall j \in \{1, 2, \cdots, N\} \\
    w^{2\rightarrow 1}_j = T_2^1(w_j^2), \   w^{1\rightarrow 2}_j = T_1^2(w_j^1), \quad  \forall j \in \{1, 2, \cdots, n\} \\
  \end{aligned}
\end{equation}
With the transformations, we can map new features to the old feature space, map old features to the new feature space or direct compare features.
These correspond to backward, forward and direct compatibility respectively.
The structure of our used transformation module is presented in top left of Fig.~\ref{fig:method1}.
The module is modified from the residual bottleneck transformation module introduced in RBT~\cite{wang2020unified}. 
Note that original module requires same feature dimension for input and output.
Thus, we append one additional fully connected layer if dimension changes.
Besides, the number of sub-modules $K=4$ for backward and forward compatibility.
For direct compatibility, $K=0$ and therefore $T_1^2, T_2^1$ degrades to identity mappings.

To train the transformations, we define four types of class regions $C_j^1, C_j^2, C_j^{1\rightarrow 2}, C_j^{2\rightarrow 1}$.
Our transformations are estimated by aligning locations of class regions  $\{ C_j^1, C_j^{2\rightarrow 1}\}$ and $\{C_j^2, C_j^{1\rightarrow 2}\}$.
In the end, the module is learned  by enforcing the consistency between class centers (\ie, $w_j^1 = w_j^{2\rightarrow 1}$ and $w_j^2 = w_j^{1\rightarrow 2}$).

\subsection{The General Framework}
Fig.~\ref{fig:method1} illustrates the details of our method. As a general framework, our method can fit both CMC and CT problems.
Specifically, model 1 is the compatible target model and therefore its weights are always fixed.
For CMC, model 2 is also fixed and the purpose is to find mappings between feature spaces of two models.
For CT, we aim at training the model 2 while keeping compatible with model 1.

\noindent
\textbf{Pre-processing.}
Before training, we first pre-compute normalized feature $f_i^1$ of training sample $\{x_i, y_i\}$ from model 1.
For each class $j \in \{1, 2, \cdots, n\}$ with $N_j$ samples, we calculate the class center $w_j^1$ by taking the average of the features.
After that, we calculate $N_j$ angles between the class center and corresponding samples and denote them as $\{\theta^1_i\}_{i=1}^{N_j}$.
Then the original boundary for class $j$ in the old feature space $b_j^1$ is estimated by
\begin{equation}
  \small
  \label{eq:boundary}
  b_j^1 = \max\{\mathcal{I}(\theta^1_1)\cdot \theta^1_1, \mathcal{I}(\theta^1_2)\cdot \theta^1_2, \cdots,  \mathcal{I}(\theta^1_{N_j})\cdot \theta^1_{N_j}\}.
\end{equation}
Here $\mathcal{I}(\theta)$ is 0 if $\theta$ is estimated as an outlier and 1 otherwise. The outliers are identified by the 1.5IQR rule which is a common method in statistical analysis~\cite{tukey1977exploratory}.

$\{w_j^2\}_{j=1}^n$ is also pre-computed for the CMC problem.
For CT, $\{w_j^2\}_{j=1}^n$ are weights in the classifier and are learned from data.
As mentioned before, our compatibility is achieved by aligning class centers between new and old models via the transformation module, and force the new-learned features to be more compact than the old ones.
That corresponds to the alignment loss $L_a$ and boundary loss $L_b$ in the framework.
A classification loss $L_{cls}$ is further introduced to guarantee model performances.

\noindent
\textbf{Alignment loss.}
Define function $d(\cdot)$ as a measurement of cosine distances.
Our alignment loss is used to align class locations between two models and defined as
\begin{equation}
  \small
  \begin{aligned}
    L_a &= \sum_{j=1}^n \left(d(w_j^{1\rightarrow 2}, w_j^2) +  d(w_j^{2\rightarrow 1}, w_j^1)\right ). 
  \end{aligned}
\end{equation}

\noindent
\textbf{Boundary loss.}
The boundary loss constrains the new features to be more centralized to class centers than the old models and is defined as
\begin{equation}
  \small
  \begin{aligned}
    L_b & = \sum_{i=1}^N\max\left(0, \theta(f_i^{2\rightarrow 1}, w_{y_i}^1) - b^1_{y_i}\right).
  \end{aligned}
\end{equation}

It not only contributes to the compatibility, but also leads to smaller intra-class variations and helps learns more discriminative embeddings.

\noindent
\textbf{Classification loss.}
The general format of a classification loss for sample $i$ is
\begin{equation}\label{eq:softmax}
  \small
    L_{cls}(w, f_i) = -\log \frac{e^{s\cdot \mu(w_{y_i}, f_i)}}{e^{s\cdot \mu(w_{y_i}, f_i)} + \overset{n}{\underset{
          j=1,
          j\neq y_i
        }{\sum}} e^{s\cdot \nu(w_{j}, f_i)}}.
  \end{equation}

$\mu, \nu$ generate similarity scores and are in various formats for different losses. For example, $\mu(w, f)=\nu(w,f)=\frac{w^Tf}{\|w^Tf\|}$ for NormFace~\cite{wang2017normface} while $\mu(w, f)=\frac{w^Tf}{\|w^Tf\|}-m, \nu(w,f)=\frac{w^Tf}{\|w^Tf\|}$ for CosFace~\cite{wang2018cosface}.

For CMC, as $f_i^1, f_i^2$ are already well-trained for classification, we compute the classification loss on $f_i^{1\rightarrow 2}$ and precalculated class centers for model 2 (\ie, $L_c = L_{cls}(w^2, f_i^{1\rightarrow 2})$).
For CT, the loss is $L_c = L_{cls}(w, f_i^2)$ where $w$ are learned during training.

\noindent
\textbf{The whole loss.}
In the end, the loss for the framework is

\begin{equation}
  \small
  \label{eq:lce}
  L_{LCE} = \lambda_aL_{a} + \lambda_bL_{b} + L_{c}
\end{equation}
Here $\lambda_a, \lambda_b$ are two hyperparameters. 

\section{Experiments}
In our experiments, we begin describing implementation details in Sec.~\ref{sec:exp_0} and evaluation metrics in Sec.~\ref{sec:exp_em}.
Sec.~\ref{sec:exp_a} is the ablation study which examines 
effects of the components in LCE.
Then we evaluate our proposed method on CMC in Sec.~\ref{sec:exp_b}.
For the CT problem, only direct compatibility is studied for fair comparisons with baselines.
Sec.~\ref{sec:exp_c} presents the results on CT with different training dataset, loss functions and network architectures, 
while experiments on different feature dimensions and sequential compatibility are in Sec.~A (in the supplementary).

\subsection{Implementation Details} \label{sec:exp_0}

\noindent
\textbf{Datasets.}
The original MS-Celeb-1M dataset~\cite{guo2016ms} contains about 10 million images of 100k identities.
However, it consists of a great many noisy face images.
Instead, MS1Mv2~\cite{Deng2018} (5.8M images, 85k identities) is adpoted as our training dataset.
For change of training datasets, we also collect faces from the first 50\% of identities from MS1Mv2 and call the dataset as MS1Mv2(1/2).
For evaluation, we adopt 
a top challenging benchmark called IJB-C~\cite{maze2018iarpa}, which covers about 3,500 identities with a total of 31,334 images and 117,542 unconstrained video frames.
We calculate the TARs at FAR=1e-4 in the 1:1 verification on IJB-C benchmark, where 19k positive and 15M negative matches are involved.
All the images are aligned to $112\times 112$ following the setting in ArcFace~\cite{Deng2018}.

\noindent
\textbf{Training.}
All the models are trained on 8 1080Tis by stochastic gradient descent.
For CMC, we train the transformations for 20 epochs, with learning rate initialized at 0.1 and divided by 10 at epoch 5, 10, 15.
For CT, we initialize the learning rate by 0.1 and divide it by 10 at 10, 18, 22 epochs.
The training stops at the 25th epoch.
The weight decay is set to 5e-4 and the momentum is 0.9.
The recommended hyper-parameters are used for classification losses from the original papers (\eg, $m=0.5, s=64$ for ArcFace~\cite{Deng2018}).
We only augment training samples by random horizontal flipping
and empirically set $\lambda_a = 100, \lambda_b = 0.1$ for LCE.

\noindent
\textbf{Baselines.}
In our experiments, the main baselines are RBT~\cite{wang2020unified} for CMC and BCT~\cite{shen2020towards} for CT, which  are current state-of-the-arts in the corresponding fields.
For RBT, we use the recommended hyper-parameters in their paper.
For BCT, the recommended weight for their influence loss are not provided.
We tried several values and found 0.5 is the best value in our implementation.
Therefore, 0.5 is used for all BCT experiments.

\subsection{Evaluation Metrics} \label{sec:exp_em}

\noindent
\textbf{Model performance and compatibility performance.}
We evaluate model performance by 1v1 verification accuracy (abbreviated as Ver. Acc. in all tables) on the IJB-C benchmark.
To measure the compatibility performance between the old model $\phi^{1}$ and the new model $\phi^{2}$, we use $\phi^{1}$ to extract the feature embedding for the first template in the pair and $\phi^{2}$ for the second, and report the cross verification accuracy (abbreviated as Cross Veri. Acc. in all tables).

\noindent
\textbf{Performance gain.}
For the scenario of CT, the performance of the upgraded model is also vital to the whole system besides the compatibility.
Therefore, we report the relative performance improvement from $\phi^{1}$ to $\phi^{2}$ by 

\begin{equation}
  \label{eq:pergain}
  \mathcal{G}_{per.} = \frac{M(\phi^{2}) - M(\phi^{1})}{|M(\phi^{2}_{upper}) - M(\phi^{1})|}.
\end{equation}

Here $M(\phi)$ measures the recognition performance of model $\phi$.
Compared to $\phi^{2}$, $\phi^{2}_{upper}$ is trained without compatibility purpose and therefore serves as the upper bound for the new model.
We add the absolute symbol to the denominator as it can be negative (\ie, $M(\phi^{2}_{upper}) < M(\phi^{1})$).
For example, the previous deployed model has a high latency and the new model is of a light-weight structure.
In this case, the model performance will degrade.

\noindent
\textbf{Upgrade gain.}
The upgrade gain is similarly defined in BCT~\cite{shen2020towards}.
We revise their definition by taking the absolute value of the denominator to handle the model degradation and define our upgrade gain as
\begin{equation}
  \label{eq:upgain}
  \mathcal{G}_{upgrade} = \frac{M(\phi^{2}, \phi^{1}) - M(\phi^{1})}{|M(\phi^{2}_{upper}) - M(\phi^{1})|}.
\end{equation}
Here $M(\phi^{2}, \phi^{1})$ measures the cross-model recognition performances between $\phi^{2}$ and $\phi^{1}$ (\ie, compatibility performances).
If using transformations, upgrade gain is the better result of backward and forward compatibility.

\subsection{Ablation Study} \label{sec:exp_a}
In this part, we conduct several ablation studies to investigate the effects of different components in our method.
ResNet50~\cite{he2016deep} is adopted as the backbone network, and we use the full version of MS1Mv2 as the training data.
The classification loss used in the old model is NormFace~\cite{wang2017normface} and that used in new models is ArcFace~\cite{Deng2018}.

{\def\arraystretch{1.2}
\setlength{\tabcolsep}{3pt}
\begin{table*}[htb]
  \begin{center}
      {\footnotesize
      \begin{tabular}{lccccccccccc}
        \hline
        New Model & Old Model   &  \multicolumn{3}{c}{Losses}  & Transformation &  Veri. & \multicolumn{3}{c}{Cross Veri. Acc.} & Perf.  & Upgrade\\
        \cline{3-5}         \cline{8-10}
                  && Alignment & Boundary & Classification & & Acc.  & Direct & Backward & Forward & Gain (\%)   & Gain (\%)\\
        \rowcolor{Gray}
        $\phi^{1}$  & -  & - & - & NormFace & - & 89.93 & - & - & - &-&-\\
        $\phi^{2}_{a}$  & $\phi^{1}$  & \cmark &  & ArcFace & & 93.63 & 92.60 & - & - & 84.28 & 60.82  \\
        $\phi^{2}_{a\_b}$  & $\phi^{1}$  & \cmark & \cmark & ArcFace &  & 93.84 & 92.93 & - & - & 89.07  & 68.34 \\
        $\phi^{2}_{a\_t}$  & $\phi^{1}$  & \cmark &  & ArcFace & \cmark &  94.16 & - &  89.25 & 92.89 & 96.36 & 67.43 \\
        $\phi^{2}_{a\_b\_t}$  & $\phi^{1}$  & \cmark & \cmark & ArcFace & \cmark  & 94.29 & - & 91.92 & 93.11 & \textbf{99.32} & \textbf{72.44} \\
        \rowcolor{Gray}
        $\phi^{2}_{upper}$  & -  & - & - & ArcFace & -  & 94.32 & 0.01 & - & - & +100.00 & -   \\
        \hline
    \end{tabular}
  }
  \end{center}
  \caption{1:1 verification TAR (\%@FAR=1e-4) on the IJB-C dataset with different combinations of components.
    } \label{table:ablation_2}
\end{table*}
}

\noindent
\textbf{Effects of components in LCE.}
We train the new models with different combinations of alignment loss, boundary loss as well as the transformation module as shown in Tab.~\ref{table:ablation_2}.
Note that $\phi^{2}_{upper}$ is trained directly without additional losses and therefore serves as the upper bound of the recognition performances for new models.
The performance gain is 84.28\% and the upgrade gain is 60.82\% with only alignment loss used.
If introducing boundary loss, the performance gain is 89.07\% and  upgrade gain increases evidently with the number of 7.52\%.
The transformation module can significantly increase the model performance as it loose the constraints of directly aligning class locations.
Performance gains of $\phi^{2}_{a\_t}$ and $\phi^{2}_{a\_b\_t}$ are all over 96\%, which is close to the upper bound.
Also, the transformation module offers the choices of whether to transform features forward and backward.
As the full version of LCE, $\phi^{2}_{a\_b\_t}$ achieves a verification accuracy of 91.92\% when transforming new features to the original feature space, and 93.11\% in the opposite direction.
In addition, $\phi^{2}_{a\_b\_t}$ achieves the best performance gain of 99.32\% and upgrade gain of 72.44\%, which demonstrates the efficacy of our proposed method.

\setlength{\tabcolsep}{1.4pt}
\begin{figure}[htb!]
  \centering
  \includegraphics[trim=0 0 0 0, clip, width=0.45\textwidth]{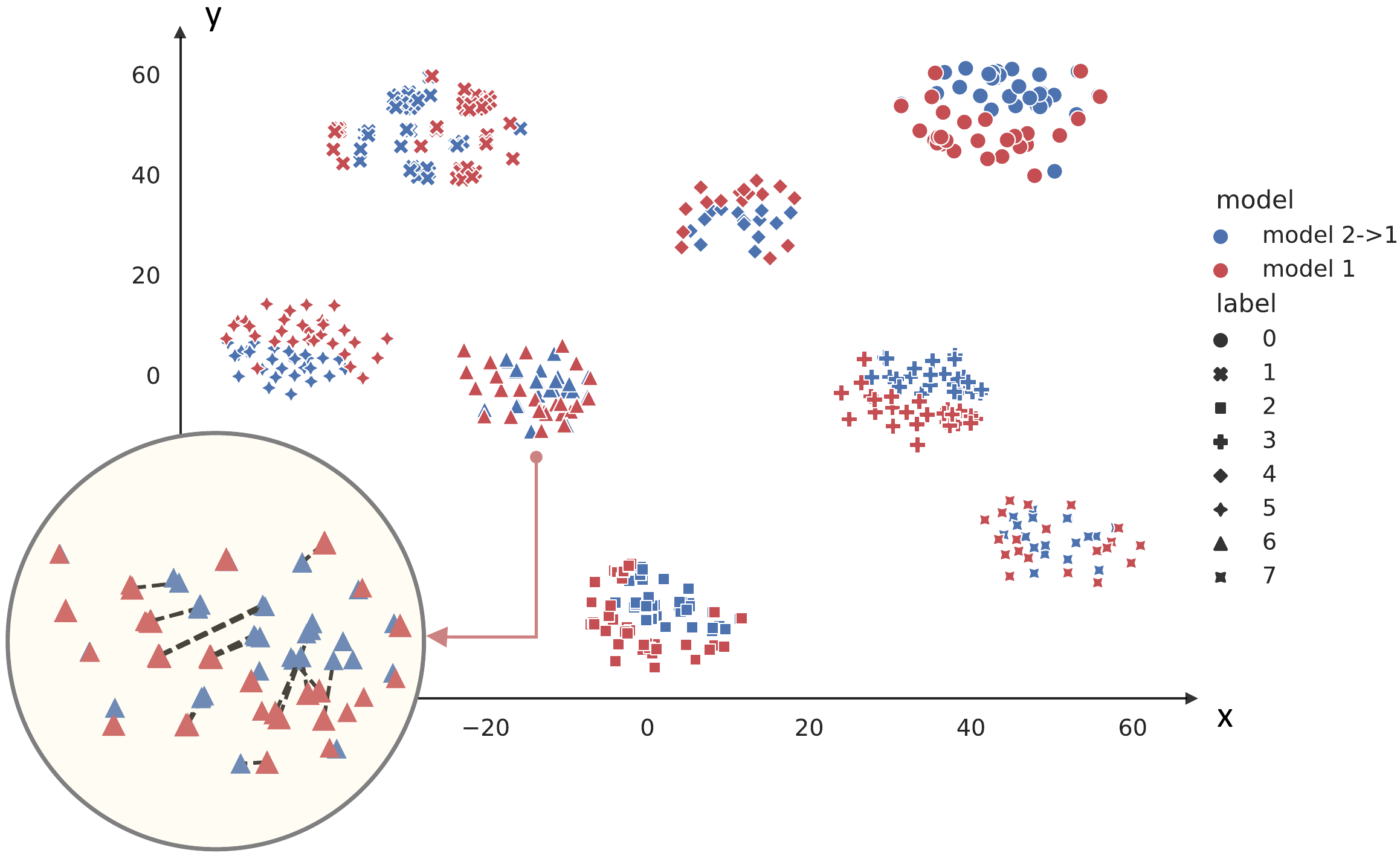}
  \caption{
    A visualization of deep features from two models.
    The area of class 6 is magnified for better revealing pair information.
    Each dash line connects features of a same instance from two models.
    \textbf{Best viewed in color}.
  }\label{fig:ablation_1}
\end{figure}

\noindent
\textbf{Visualization.}
To reveal the effects of the proposed method, we sample 8 classes from MS1Mv2 and visualize deep features in Fig.~\ref{fig:ablation_1}.
Here model 1 is $\phi^{1}$ and model 2 is  $\phi^{2}_{a\_b\_t}$ from Tab.~\ref{table:ablation_2}.
Model 2$\rightarrow$1 produces features by mapping those from model 2 to the feature space of model 1 by the learned transformations.
T-SNE is chosen to map high-dimensional features into top 2 dimensions.
The figure demonstrates that the mapped features are well-aligned with the original features and have more compact intra-class distributions than original ones.

We further pick class 6 and visualize feature pairs of its first 10 samples to show their pair-wise relationships.
Features of the same image from two models can be either close to or far away from each other, as shown in the figure, which reveals our compatibility works in a point-to-set instead of strict point-wise manner.

Tab.~\ref{table:ablation_3} presents the intra-class and inter-class distances of three types of features.
The average intra-class distance is calculated by $\frac{1}{N} \sum_{i=1}^N (\frac{f_i}{\|f_i\|} - \frac{w_{y_i}}{\|w_{y_i}\|})^2$ and the average inter-class distance is calculated by $\frac{1}{n(n-1)}\sum_{i}^n\sum_{j=i+1}^n(\frac{w_i}{\|w_i\|} - \frac{w_j}{\|w_j\|})^2$.
Because of the proposed alignment and boundary loss,  model 2$\rightarrow $1 shrinks the distribution of each class and therefore its features have much smaller intra-class distances than the original model 1 (0.283 \vs 0.347).
The transformations further add flexibility of class locations and helps model to learn more separated inter-class distributions.
The average inter-class distance increases from 1.767 to 1.827 in this case.

\setlength{\tabcolsep}{3pt}
\begin{table}
  \begin{center}
    {\footnotesize
      \begin{tabular}{lcccccc}
        \hline
         & model 1 & model 2$\rightarrow $1 & model 2\\
        \hline
        intra-class  & 0.347 & 0.283 & 0.277 \\
        inter-class  & 1.767 & 1.792 & 1.827        \\
        \hline
    \end{tabular}
  }
  \end{center}
  \caption{Average intra-class and inter-class distances of features from different models.} \label{table:ablation_3}
\end{table}

\subsection{Cross Model Compatibility} \label{sec:exp_b}
\setlength{\tabcolsep}{3pt}
\begin{table}
  \begin{center}
      {\footnotesize
      \begin{tabular}{cccccc}
        \hline
        Model & Backbone & Loss  & Training Dataset & Ver. Acc.\\
        \hline
        $\phi^*_a$ & ResNet100 & ArcFace~\cite{Deng2018} & MS1Mv2 & 95.72 \\
        $\phi^*_b$ & ResNet50 & ArcFace~\cite{Deng2018} & MS1M-REFINE-v1 & 92.52 \\
        $\phi^*_c$ & ResNet50 & Subcenter~\cite{deng2020subcenter} & MS1Mv0 & 95.44 \\
        \hline
    \end{tabular}
  }
  \end{center}
  \caption{Details of selected pretrained models.
    Here MS1Mv0 is the original MS-Celeb-1M dataset~\cite{guo2016ms} while the other two datasets are cleaned versions by authors of ArcFace~\cite{Deng2018}.
    Verification accuracies (\%) are TARs@FAR=1e-4 on IJB-C. } \label{table:exp_b1}
\end{table}

For CMC, we adopt three pretrained models provided by the InsightFace project\footnote{https://github.com/deepinsight/insightface/}, which are of different network architectures and trained on different version of MS1M datasets as well as different losses.
Details of the models and their abbreviations can be found in Tab.~\ref{table:exp_b1}.
We select NormFace~\cite{wang2017normface} and ArcFace~\cite{Deng2018} as the classification loss to train the transformations and present the experimental results in Tab.~\ref{table:exp1_1}.

RBT maps features from two models to a third feature space.
In contrast, our LCE maps feature forward or backward and therefore has two cross verification accuracies in each experiment.
In all results, $(\phi_a^*, \phi_c^*)$ always have the best cross verification accuracies than other two combinations.
That indicates compatibility performances are also influenced by performances of source models.
When using ArcFace, both methods achieve similar compatibility performances.
While using NormFace, the performances of RBT drop dramatically with more than 1\% for all cases.
That is because RBT compare features in the third feature space and therefore the performance is heavily relied on the goodness of that feature space.
In contrast, our method is robust to changes of classification losses and achieves consistent high cross verification accuracies.
Besides the robustness, the computational costs of LCE are only half of RBT.
Our method is also more flexible in real-world applications as one can choose to either re-encode the feature in gallery sets or map new feature to the old feature space.

\setlength{\tabcolsep}{3pt}
\begin{table}
  \begin{center}
      {\footnotesize
      \begin{tabular}{cccccc}
        \hline
        \multirow{2}{*}{Method}  & Classification & \multirow{2}{*}{$(\phi^*_{a}, \phi^*_{b})$} &   \multirow{2}{*}{$(\phi^*_{a}, \phi^*_{c})$} & \multirow{2}{*}{$(\phi^*_{b}, \phi^*_{c})$}   \\
        & Loss\\
        \rowcolor{Gray}
        direct & -  & 0.01 & 0.01 & 0.03 \\
        RBT~\cite{wang2020unified} & ArcFace & 94.04 & \textbf{95.29} & 93.68 \\
        LCE  & ArcFace & \textbf{94.08}/93.90 & 95.19/95.27 & \textbf{93.72}/93.51\\
        \hline
        RBT~\cite{wang2020unified} & NormFace & 92.78 & 94.01 & 92.33 \\
        LCE  & NormFace & \textbf{94.08}/93.86 &  \textbf{95.20}/95.02 & \textbf{93.72}/93.40 \\
        \hline
    \end{tabular}
  }
  \end{center}
  \caption{1:1 verification TAR (\%@FAR=1e-4) on the IJB-C for cross model compatibility.} \label{table:exp1_1}
\end{table}

\subsection{Compatible Training} \label{sec:exp_c}
In this section and Sec.~A in the supplementary, we explore the scenario of CT.
The new models are upgraded from a compatible target model with the change of different training datasets, classification losses, network architectures as well as the feature dimensions.
We adopt two baselines:
(1) a naive KD baseline which uses $l^2$-distance and
(2) BCT~\cite{shen2020towards}.
For fair comparison with BCT~\cite{shen2020towards}, we set $K=0$ in the transformation module and only consider direct compatibility.
We call the old model as $\phi^{1}$, the new model without compatibility losses as $\phi^{2}_{upper}$  and with $l_2$ baseline, BCT, LCE as $\phi^{2}_{l2}, \phi^{2}_{bct}, \phi^{2}_{lce}$, respectively.
In all the experiments below, cross verification accuracies between $\phi^{1}$ and $\phi^{2}_{upper}$ are all near 0, which demonstrates that model compatibility cannot be directly achieved without usage of compatibility methods.

Unless stated otherwise, the default setting is to employ MS1Mv2 as the training dataset, ResNet50~\cite{he2016deep} as the backbone network, ArcFace as the classification loss and feature dimension as 512.
Below we present results with one factor changed each time while others are set to default.
Note that $\phi^{1}$ in different tables may refer to different models.


\noindent
\textbf{Changes of training datasets.}
The old model $\phi^{1}$ is trained on MS1Mv2(1/2) and all the new models are trained with the full version of MS1Mv2.
Tab.~\ref{table:lce_dataset} presents the results.
$\phi^{2}_{l_2}$ directly matches features and therefore has a performance similar to the previous one.
$\phi^{2}_{bct}$ has a performance gain of 47.06\% but its upgrade gain is close to $\phi^{2}_{l_2}$ (both around 23\%).
A possible reason is that the old model is trained on a reduced dataset and therefore cannot clearly separate classes in the full data.
The point-wise compatibility constraints used in baselines prevent simultaneous improving recognition performances and achieving model compatibility.
In contrast, we use a point-to-set constraint and that makes our $\phi^{2}_{lce}$ achieve much better results.
The gains are 84.31\% and 81.37\% respectively in our implementation.

{\def\arraystretch{1.2}
\setlength{\tabcolsep}{3pt}
\begin{table}[htb]
  \begin{center}
    {\footnotesize
      \begin{tabular}{lcccc}
        \hline
        \multirow{2}{*}{Model} & Veri. & Cross Veri. & Perf.  & Upgrade\\
        & Acc.  & Acc. & Gain (\%)   & Gain (\%)\\
        \rowcolor{Gray}
        $\phi^{1}$ & 93.16  & - & - & - \\
        $\phi^{2}_{l2}$ & 92.92 &  93.40 & -23.53 & +23.53  \\
        $\phi^{2}_{bct}$ & 93.64 & 93.39 & +47.06 & +22.54 \\
        $\phi^{2}_{lce}$ & 94.02 & 93.99 & \textbf{+84.31} & \textbf{+81.37}  \\
        \rowcolor{Gray1}
        $\phi^{2}_{upper}$ & 94.18 & 0.02  & +100.00 & - \\
        \hline
    \end{tabular}
  }
  \end{center}
  \caption{1:1 verification TAR (\%@FAR=1e-4) on the IJB-C dataset with changes of training datasets. 
  } \label{table:lce_dataset}
\end{table}
}

\noindent
\textbf{Changes of classification losses.}
With old model trained on ArcFace,  we train new models on different classification losses.
We pick NormFace~\cite{wang2017normface}, CosFace~\cite{wang2018cosface} and report the results in Tab.~\ref{table:lce_loss}.
The old model is trained on the best classification loss and therefore can separate classes well.
$\phi^{2}_{l2}$ trained by the naive approach performs poorly in this scenario.
Our $\phi^{2}_{lce}$ still achieve better results especially for the model compatibility.
In particular, the upgrade gains are 27.69\% and 12.76\% higher than those from $\phi^{2}_{bct}$ when training with CosFace.

{\def\arraystretch{1.2}
\setlength{\tabcolsep}{3pt}
\begin{table}[htb]
  \begin{center}
    {\footnotesize
      \begin{tabular}{lccccc}
        \hline
        \multirow{2}{*}{Model} & Classification & Veri. & Cross Veri. & Perf.  & Upgrade\\
                               & Loss & Acc.  & Acc. &  Gain (\%) &  Gain (\%) \\
        \rowcolor{Gray}
        $\phi^{1}$ & ArcFace & 94.18  & - & - & - \\
        \hline
        $\phi^{2}_{l2}$ & CosFace &  92.79 & 93.09 & -106.92 & -83.84 \\
        $\phi^{2}_{bct}$ & CosFace & 93.56 & 93.20 & -47.69 & -75.38 \\
        $\phi^{2}_{lce}$ & CosFace & 93.53  & 93.56 & \textbf{-50.00} & \textbf{-47.69} \\
        \rowcolor{Gray1}
        $\phi^{2}_{upper}$ & CosFace & 92.88 & 0.01 & -100.00 & - \\
        \hline
        $\phi^{2}_{l2}$ & NormFace &  89.28 & 92.32 & -192.91 & -73.22 \\
        $\phi^{2}_{bct}$ & NormFace & 90.42 & 93.25 & -97.92 & -24.22  \\
        $\phi^{2}_{lce}$ & NormFace & 90.58 & 93.74 & \textbf{-93.75} & \textbf{-11.46} \\
        \rowcolor{Gray1}
        $\phi^{2}_{upper}$ & NormFace & 90.34  & 0.00 & -100.00 & - \\
        \hline
    \end{tabular}
  }
  \end{center}
  \caption{1:1 verification TAR (@FAR=1e-4) on the IJB-C dataset with changes of classification losses.} \label{table:lce_loss}
\end{table}
}

Another interesting phenomenon is that the verification performances of $\phi^{2}_{lce}, \phi^{2}_{bct}$ are higher than the $\phi^{2}_{upper}$.
This is probably because the old model performs better than $\phi^{2}_{upper}$, the compatibility losses in $\phi^{2}_{lce}, \phi^{2}_{bct}$ help transferring good knowledge to new models.
In Tab.~\ref{table:lce_backbone}, the old model is also better than $\phi^{2}_{upper}$ with  ResNet50 when new models have structures of ResNet18 or MobileFace.
However, we do not observe similar phenomenon.
We guess the reason is that compared to losses, knowledge is hard to transfer across models with different network architectures.

\noindent
\textbf{Changes of network architectures.}
In this part, we study the model compatibility across different network architectures.
The old model is trained with ResNet50 while the new models are trained with ResNet100, ResNet18~\cite{he2016deep} or MobileFaceNet~\cite{chen2018mobilefacenets}.
Results are presented in Tab.~\ref{table:lce_backbone}.
In all cases, $\phi^{2}_{l2}$ has good performance gains while poor upgrade gains.
The gains of $\phi^{2}_{bct}$  are in completely opposite patterns.
In contrast, our $\phi^{2}_{lce}$ achieves remarkable results for both recognition and compatibility performance, which demonstrates the superiority of the proposed method.

{\def\arraystretch{1.2}
\setlength{\tabcolsep}{3pt}
\begin{table}
  \begin{center}
    {\footnotesize
      \begin{tabular}{lccccc}
        \hline
        \multirow{2}{*}{Model} & \multirow{2}{*}{Backbone} & Veri. & Cross Veri. & Perf.  & Upgrade\\
        && Acc.  & Acc. &  Gain (\%) &  Gain (\%) \\
        \rowcolor{Gray}
        $\phi^{1}$ & ResNet50 & 94.18  & - & - & -\\
        \hline
        $\phi^{2}_{l2}$ & ResNet18 &  89.52 &  90.10 & -119.79 & -104.88  \\
        $\phi^{2}_{bct}$ & ResNet18 & 88.78 & 92.06 & -138.82 & -54.49 \\
        $\phi^{2}_{lce}$ & ResNet18 & 89.71 & 92.81 & \textbf{-114.91} & \textbf{-35.22} \\
        \rowcolor{Gray1}
        $\phi^{2}_{upper}$ & ResNet18 & 90.29 & 0.01 & -100.00 & - \\
        \hline
        $\phi^{2}_{l2}$ & MobileFace &  87.60 & 83.93 & -109.67 & -170.83\\
        $\phi^{2}_{bct}$ & MobileFace & 87.10 & 91.46 & -118.00 & \textbf{-45.33} \\
        $\phi^{2}_{lce}$ & MobileFace & 87.84 & 91.31 & \textbf{-105.66} & -47.50 \\
        \rowcolor{Gray1}
        $\phi^{2}_{upper}$ & MobileFace & 88.18  & 0.00 & -100.00 & - \\
        \hline
        $\phi^{2}_{l2}$ & ResNet100 & 94.40  & 93.43 & +22.45 & -76.53 \\
        $\phi^{2}_{bct}$ & ResNet100 & 94.01 & 94.69 & -17.35 & +52.04\\
        $\phi^{2}_{lce}$ & ResNet100 & 94.64 & 95.07 & \textbf{+46.94} & \textbf{+90.82}\\
        \rowcolor{Gray1}
        $\phi^{2}_{upper}$ & ResNet100 & 95.16 & 0.03 & +100.00 & - \\
        \hline
    \end{tabular}
  }
  \end{center}
  \caption{1:1 verification TAR (\%@FAR=1e-4) on the IJB-C dataset with changes of network architectures.} \label{table:lce_backbone}
\end{table}
}

\section{Conclusion}
In this paper, we design a general framework called LCE for model compatibility.
Our compatibility works in a point-to-set manner which is realized by aligning the class centers between models and restricting more compact intra-class distributions for the new model.
The adequate experimental results demonstrate that our method can achieve model compatibility in various scenarios while with marginal sacrifices of accuracies.
As a general framework, our method can be applied to different combinations of scenarios (CMC/CT) and compatible directions (direct/forward/backward).
Moreover, it can be potentially extended to benefit other tasks such as person/car re-identification and image retrieval.

\clearpage

{\small
\bibliographystyle{ieee_fullname}
\bibliography{egbib}
}

\clearpage
\appendix
\section{Compatible Training} \label{sec:exp_c}

\noindent
\textbf{Changes of feature dimensions.}

{\def\arraystretch{1.2}
\setlength{\tabcolsep}{1pt}
\begin{table}[htb!]
  \begin{center}
    \resizebox{0.48\textwidth}{!}{%
      \begin{tabular}{lccccccc}
        \hline
        \multirow{2}{*}{Model} & Feature & Veri. & \multicolumn{3}{c}{Cross Veri. Acc.} & Perf.  & Upgrade\\
        \cline{4-6}        
                               & Dimension & Acc. & Direct & Backward&  Froward &  Gain (\%)&  Gain (\%) \\
        \rowcolor{Gray}        
        $\phi^1$ & 256 & 93.99  & - & - & - & & \\
        \hline
        $\phi^2_{l2}$ & 512 & 93.94 & - & 93.53 & 93.85 & -26.32 & -73.68 \\
        $\phi^2_{bct}$ & 512 & 93.58 & 93.77 & - & - & -215.79 & -115.79 \\
        $\phi^2_{lce}$ & 512 & 94.01 & -  & 91.85 & 94.27 & +10.53 & +147.37 \\
        \rowcolor{Gray1}                
        $\phi^2_{upper}$ & 512 & 94.18 & - & - & -& +100.00 & - \\        
        \hline        
    \end{tabular}
  }    
  \end{center}
  \caption{1:1 verification TAR (\%@FAR=1e-4) on the IJB-C dataset~\cite{maze2018iarpa} with increasing feature dimensions.} \label{table:lce_featdim}
\end{table}
}

{\def\arraystretch{1.2}
\setlength{\tabcolsep}{3pt}
\begin{table}[htb!]
  \begin{center}
    {\footnotesize
      \begin{tabular}{lcccccc}
        \hline
        \multirow{2}{*}{Model} & Feature & Veri. & \multicolumn{2}{c}{Cross Veri. Acc.} & Perf.  & Upgrade\\
        \cline{4-5}        
                               & Dimension & Acc.  & Backward&  Froward &  Gain (\%)&  Gain (\%) \\
        \rowcolor{Gray}        
        $\phi^1$ & 512 & 94.18  & - & - & - & \\
        \hline
        $\phi^2_{l2}$ & 256 & 90.49 & 91.59 & 90.29 & -1942.11 & -1363.16 \\
        $\phi^2_{lce}$ & 256 & 93.69 & 92.40 & 94.02 & -257.89 & -84.21 \\
        \rowcolor{Gray1}                
        $\phi^2_{upper}$ & 256 & 93.99 & - & -& -100.00 & - \\        
        \hline
    \end{tabular}
  }    
  \end{center}
  \caption{1:1 verification TAR (\%@FAR=1e-4) on the IJB-C dataset~\cite{maze2018iarpa} with decreasing feature dimensions.} \label{table:lce_featdimdec}
\end{table}
}


Changes of feature dimensions can be applied on increasing or decreasing feature dimensions from the old to the new model. 
For the scenario of increasing feature dimensions, 256 and 512 are used as the feature dimensions for the old and the new model, respectively.
We reverse the feature dimensions of the old and the new model when experimenting decreasing feature dimensions.
Tab.~\ref{table:lce_featdim} represents the results of dimension increasing, where our proposed LCE framework $\phi^2_{lce}$ is compared with $\phi^2_{bct}$ and ${\phi^2_{l2}}$.
$\phi^2_{l2}$ acts negatively on performance and upgrade gains, and $\phi^2_{bct}$ performs even worse on both criterions.
Compared to them, our approach $\phi^2_{lce}$ earns a much higher upgrade gain while persisting a positive performance gain.

Similar results are represented in Tab.~\ref{table:lce_featdimdec} for dimension decreasing.
$\phi^2_{l2}$ ends up with catastrophic scores on performance and upgrade gains. In contrast, $\phi^2_{lce}$ presents considerable superiority on both criterions.
Since $\phi^2_{bct}$ is not capable of dimension decreasing, results of $\phi^2_{bct}$ are excluded in Tab.~\ref{table:lce_featdimdec}. 
This further emphasizes the flexibility of our LCE framework that is capable of both dimension increasing and decreasing scenarios.

\noindent
\textbf{Multi-model and sequential compatibility.}

{\def\arraystretch{1.2}
\setlength{\tabcolsep}{3pt}
\begin{table}[htb!]
  \begin{center}
    {\footnotesize
      \begin{tabular}{c|ccc}
        \hline
        \rowcolor{Gray1}        
        & $\phi^1$ & $\phi^2$ & $\phi^3$ \\
        \hline
        \cellcolor{Gray1}
        $\phi^1$ & 91.00 & 91.80 & 91.87 \\
        \cellcolor{Gray1}
        $\phi^2$ & - & 93.05 & 93.65 \\
        \cellcolor{Gray1}
        $\phi^3$ & - & - & 94.30 \\
        \hline        
    \end{tabular}
  }    
  \end{center}
  \caption{1:1 verification TAR (\%@FAR=1e-4) on the IJB-C dataset~\cite{maze2018iarpa} with sequential changes on training datasets.} \label{table:lce_seqcom}
\end{table}
}

Multi-model and sequential compatibility is utilized to where three or more different models are required to be compatible with each other, which is commonly exists in industrial scenarios such as performing sequential model upgrades. 
To verify sequential compatibility, three versions of models $\phi^1$, $\phi^2$, $\phi^3$ are trained with 25\%, 50\%, 100\% identities from MS1Mv2~\cite{Deng2018} dataset, respectively. 
$\phi^1$ is viewed as the initial version and thus trained without compatibility constraints. 
We endow $\phi^2$ with LCE constraints that guarantee compatibility with $\phi^1$, and $\phi^3$ with LCE constraints that guarantee compatibility with $\phi^2$.
Self-verifications are implemented on $\phi^1$, $\phi^2$ and $\phi^3$ themselves, whose results are considered as lower/upper bound for cross-model verifications.
Cross-model verifications are performed between all possible permutations of model pairs from $\phi^1$, $\phi^2$ and $\phi^3$. 
Results are represented in Tab.~\ref{table:lce_seqcom}. 
Each TAR of cross-model stays between the lower and upper bound from its model pair, which indicates that $\phi^1$, $\phi^2$ and $\phi^3$ are compatible with each other.

\noindent
\textbf{Transformation module and compatible directions.}

{\def\arraystretch{1.2}
\setlength{\tabcolsep}{3pt}
\begin{table*}[htb]
  \begin{center}
    {\footnotesize
      \begin{tabular}{lcccccccc}
        \hline
        \multirow{2}{*}{Model} & \multirow{2}{*}{Backbone} & \multirow{2}{*}{Transformation} & Veri. & \multicolumn{3}{c}{Cross Veri. Acc.} & Perf. & Upgrade \\
        \cline{5-7}
        &&& Acc. & Direct & Backward&  Forward &  Gain (\%)&  Gain (\%) \\
        \rowcolor{Gray}        
        $\phi^1$ & ResNet50 & - & 94.18 & - & - & - & - & - \\
        \hline
        $\phi_{lce}^2$ & ResNet18 & & 89.71 & 92.81 & - & - & -114.91 & -35.22 \\
        $\phi_{lce\_t}^2$ & ResNet18 & \cmark & 90.46 & - & 92.09 & 92.76 & -95.63 & -36.50 \\
        \rowcolor{Gray1}                
        $\phi_{upper}^2$ & ResNet18 & - & 90.29 & 0.01 & - & -& -100.00 & - \\
        \hline
        $\phi_{lce}^2$ & MobileFace & & 87.84 & 91.31 & - & - & -105.66 & -47.50 \\
        $\phi_{lce\_t}^2$ & MobileFace & \cmark & 88.83 & - & 89.76 & 91.80 & -89.17 & -39.67 \\
        \rowcolor{Gray1}
        $\phi_{upper}^2$ & MobileFace & - & 88.18 & 0.00 & - & -& -100.00 & - \\
        \hline
        $\phi_{lce}^2$ & ResNet100 & & 94.64 & 95.07 & - & - & +46.94 & +90.82 \\
        $\phi_{lce\_t}^2$ & ResNet100 & \cmark & 94.87 & - & 94.76 & 95.04 & +70.41 & +87.76 \\
        \rowcolor{Gray1}
        $\phi_{upper}^2$ & ResNet100 & - & 95.16 & 0.03 & - & -& +100.00 & - \\
        \hline
    \end{tabular}
  }    
  \end{center}
  \caption{1:1 verification TAR (\%@FAR=1e-4) on the IJB-C dataset~\cite{maze2018iarpa} with different compatible directions.} \label{table:lce_trans}
\end{table*}
}

In this section we extend Tab. 7 of Sec. 4 with transformation module introduced during LCE training, aiming at verifying the effectiveness of the transformation module as well as model compatibility for each compatible direction.
Upgraded models trained with both LCE constraints and the transformation module are noted as $\phi^2_{lce\_t}$.
Results of introducing transformation module are summarized in Tab.~\ref{table:lce_trans} where the performance gain of each $\phi^2_{lce\_t}$ witnesses an increment of about 20\% compared with $\phi^2_{lce}$, which infers a positive impact. 

Tab.~\ref{table:lce_trans} also reveals model compatibility achievements of direct, backward and forward compatible directions.
Observed from each backbone, cross verification accuracy of the backward direction performs slightly worse than the direct compatible method but is still considerable, while the forward compatible method has a competitive (or even better) cross verification accuracy compared with the direct manner.
This indicates that our framework guarantees model compatibility for each compatible direction.

\section{Experiments on Person Re-Identification}

Besides the face recognition, we also conduct experiments on person re-identification to further reveal the generalization of our method.
Our implementation is based on a public repository\footnote{https://github.com/guxinqian/Simple-ReID} provided by authors of Gu \etal~\cite{gu2020AP3D}.
We use the market-1501 dataset~\cite{Zheng2015Scalable} consisting of  1501 identities and 32217 images.
Following the protocol in Zheng \etal~\cite{Zheng2015Scalable}, 751 identities are reserved for training and the remaining 750 identities are used for testing.
We use the first half of identities to train the old model and the full dataset to train the new model.
The results are reported in Tab.~\ref{tab:reid}.

\begin{table}[htb!]
  \centering
  \setlength{\tabcolsep}{3pt}
  \footnotesize      
  \begin{tabular}{c|ccc|cc}
    \hline
    Model & $\phi^1$ & $\phi^2_{upper}$ & $\phi^2_{lce}$ & $(\phi^1, \phi^2_{upper})$ &  $(\phi^1, \phi^2_{lce})$\\
    \hline
    mAP & 76.9 & 86.2 & 86.1 & 70.0 & 77.4 (+7.4) \\
    \hline                                  
  \end{tabular}
  \caption{Mean average percision (mAP) (\%) on Market-1501.}  \label{tab:reid}  
\end{table}

The old model $\phi^1$ achieves a mAP of 76.9\% and the upper bound model gets 86.2\%.
If using LCE, the performance degrades an acceptable value of  0.1\%.
For the compatibility, our model achieves 77.4\%, which is 7.4\% better than the model trained without compatibility constraints.
The results have demonstrated the efficacy of our method.

\section{Visualization}



\begin{figure}
  \centering
  \subfloat[\label{fig:vis_l2}]{
    \includegraphics[trim=95 60 77 58,clip,width=0.23\textwidth]{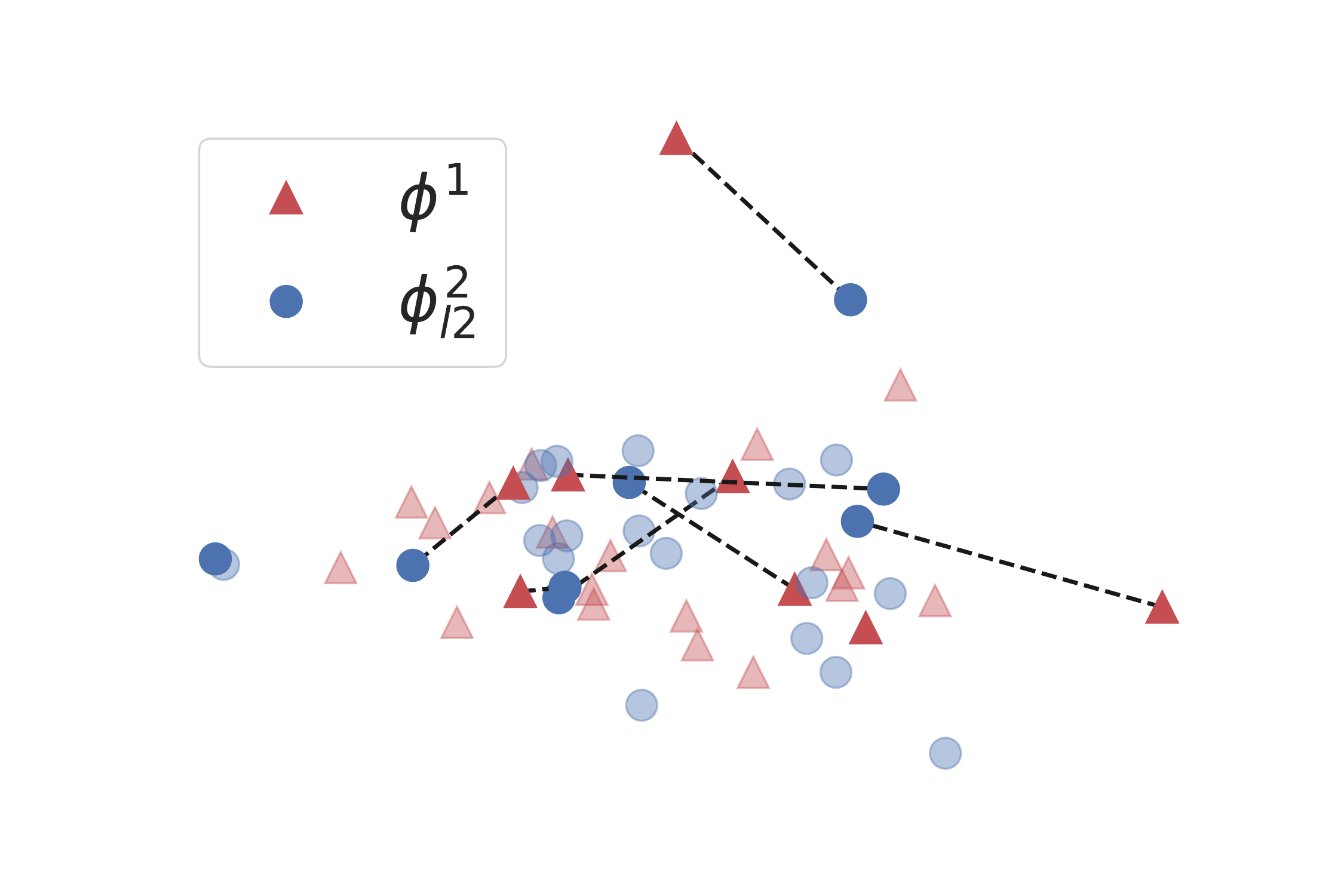}}
  \subfloat[\label{fig:vis_lce}]{
    \includegraphics[trim=95 60 77 58,clip,width=0.23\textwidth]{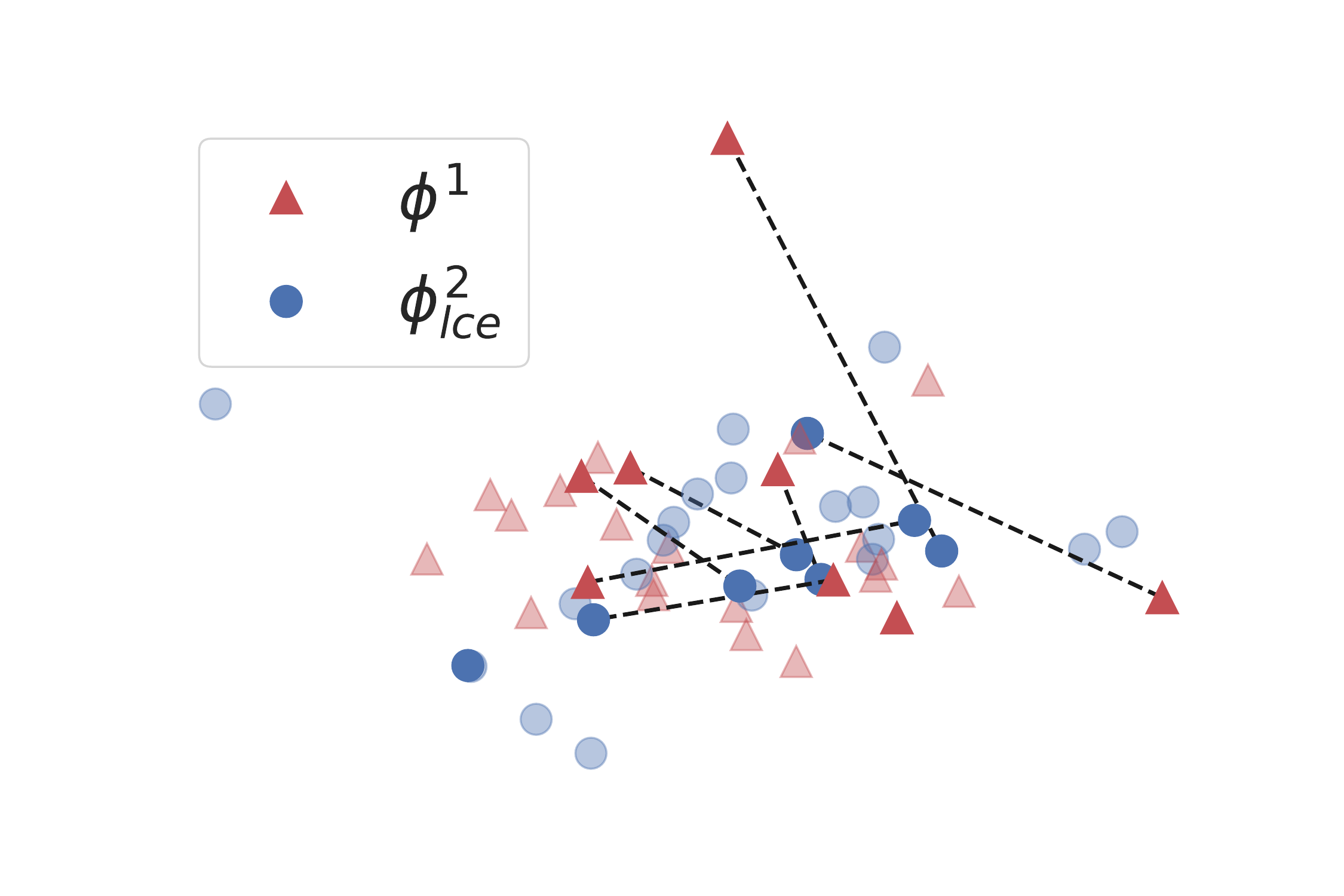}} 
  \caption{Visualizations of deep features from a same class.
  (a) Deep features from $\phi^1$ and $\phi^1_{l_2}$. The average pair-wise distance $d_{pw}(\phi^2_{l_2},\phi^1)$ is 0.230.
  (b) Deep features from $\phi^1$ and $\phi^2_{lce}$. The average pair-wise distance $d_{pw}(\phi^2_{lce},\phi^1)$ is 0.409.
  \textbf{Best viewed in color}.}
  \label{fig:vis}
\end{figure}

{\def\arraystretch{1.2}
\setlength{\tabcolsep}{3pt}
\begin{table}[htb!]
    \begin{center}
      {\footnotesize
        \begin{tabular}{lcccc}
          \hline
          & $\phi^1$ & $\phi^2_{l2}$ & $\phi^2_{lce}$ \\
          
          \hline
          intra-class  & 0.437 & 0.362 & 0.344 \\
          \hline
      \end{tabular}
    }
    \caption{Average intra-class distances of features from different models.} \label{table:vis_1}    
    \end{center}
          
\end{table}
}

        

To further study the effects of our LCE constraints, we sample 8 classes from MS1Mv2~\cite{Deng2018} and visualize one of the classes in Fig.~\ref{fig:vis}.
Models used to extract those features are chosen from Tab. 5 of Sec. 4.5 where LCE is conducted in the direct compatible method.
For Fig.~\ref{fig:vis_l2}, the new model is $\phi^2_{l_2}$ and serves as a baseline.
For Fig.~\ref{fig:vis_lce}, we use $\phi^2_{lce}$ as the new model.
$\phi^1$ is the old model for both two figures.
Average intra-class distances are calculated upon the 8 classes as Sec. 4.3 mentioned. 
Tab.~\ref{table:vis_1} represents the intra-class distances of three types of features where $\phi^2_{lce}$ produces a smaller intra-class distance than $\phi^2_{l2}$ and the original old model $\phi^1$, and this shows the capability of our LCE framework to shrink distributions of features in the same class.

A new metric called average pair-wise distance $d_{pw}(\cdot)$ is introduced to measure the expected distance between each feature pair of the 8 classes by calculating $d_{pw}(\phi^1,\phi^2)=\frac{1}{N}\sum_{i=1}^N (\frac{f^1_i}{\|f^1_i\|}-\frac{f^2_i}{\|f^2_i\|})^2$, where $f^1_i$, $f^2_i$ represents the $i^{th}$ feature pair from model $\phi^1$, $\phi^2$ respectively.
Fig.~\ref{fig:vis} presents the average pair-wise distances of two types of feature pairs where $d_{pw}(\phi^2_{lce},\phi^1)$ has a greater value than $d_{pw}(\phi^2_{l2},\phi^1)$. The results indicate that our method works in a point-to-set scheme that provides flexibility for feature locations.

\end{document}